\pdfoutput=1

\documentclass[11pt]{article}

\usepackage[final]{naacl2027}
\usepackage{times}
\usepackage{latexsym}
\usepackage[T1]{fontenc}
\usepackage[utf8]{inputenc}
\usepackage{microtype}
\usepackage{inconsolata}
\usepackage{booktabs}
\usepackage{graphicx}
\usepackage{xspace}
\usepackage{amsmath}
\usepackage{algorithm}
\usepackage{algpseudocode}
\usepackage{amssymb}
\usepackage{multirow}
\usepackage{enumitem}
\usepackage{xcolor}
\usepackage{pifont}
\usepackage{tikz}
\usetikzlibrary{positioning,arrows.meta,calc}

\newcommand{\name}{\textsc{Murmur}\xspace}

\title{\name: Efficient Long-Context ASR Inference with Hierarchical Chunking}

\author{Wei-Tzu Lee \quad Keisuke Kamahori \quad Baris Kasikci \\
  University of Washington \\
  \texttt{\{rubywtl, kamahori, barisk\}@uw.edu}}

\begin{document}
\pagestyle{plain}
\maketitle
\thispagestyle{plain}

\begin{abstract}
Long-form automatic speech recognition (ASR) requires both high accuracy and low inference latency, but existing systems fall short of handling the trade-off between the two.
Chunk-based approaches process audio in parallel windows for low latency, but lose cross-chunk context and need brittle heuristics to align speakers and timestamps at boundaries.
Long-context ASR models transcribe everything in a single pass for better accuracy, but run several times slower, and a single repetition loop can corrupt the entire output.
In this paper, we first analyze how context length affects the accuracy of long-context ASR models and find that an intermediate chunk size of about five minutes matches single-pass accuracy while avoiding repetition loops.
Building on this, we propose \name{}, an inference system for long-form ASR based on hierarchical chunking.
The long-context model decodes intermediate target chunks in parallel, and a fast short-window ASR on 30\,s draft windows accelerates each chunk with lossless, structure-aware speculative decoding.
The short-window ASR drafts the transcript, simple format rules draft the timestamps and speaker tags, and the long-context model verifies every token.
On two long-context ASR models and four meeting datasets in English and Mandarin, \name cuts end-to-end latency per recording by 2.5$\times$ on average (1.9--2.9$\times$) relative to single-pass decoding, including diarization, while lowering tcpWER by 1.9 points on average.
Our code is available at \url{https://github.com/uw-syfi/Murmur}.

\end{abstract}

\section{Introduction}
\label{sec:intro}

\begin{figure}[t]
    \centering
    \includegraphics[width=\linewidth]{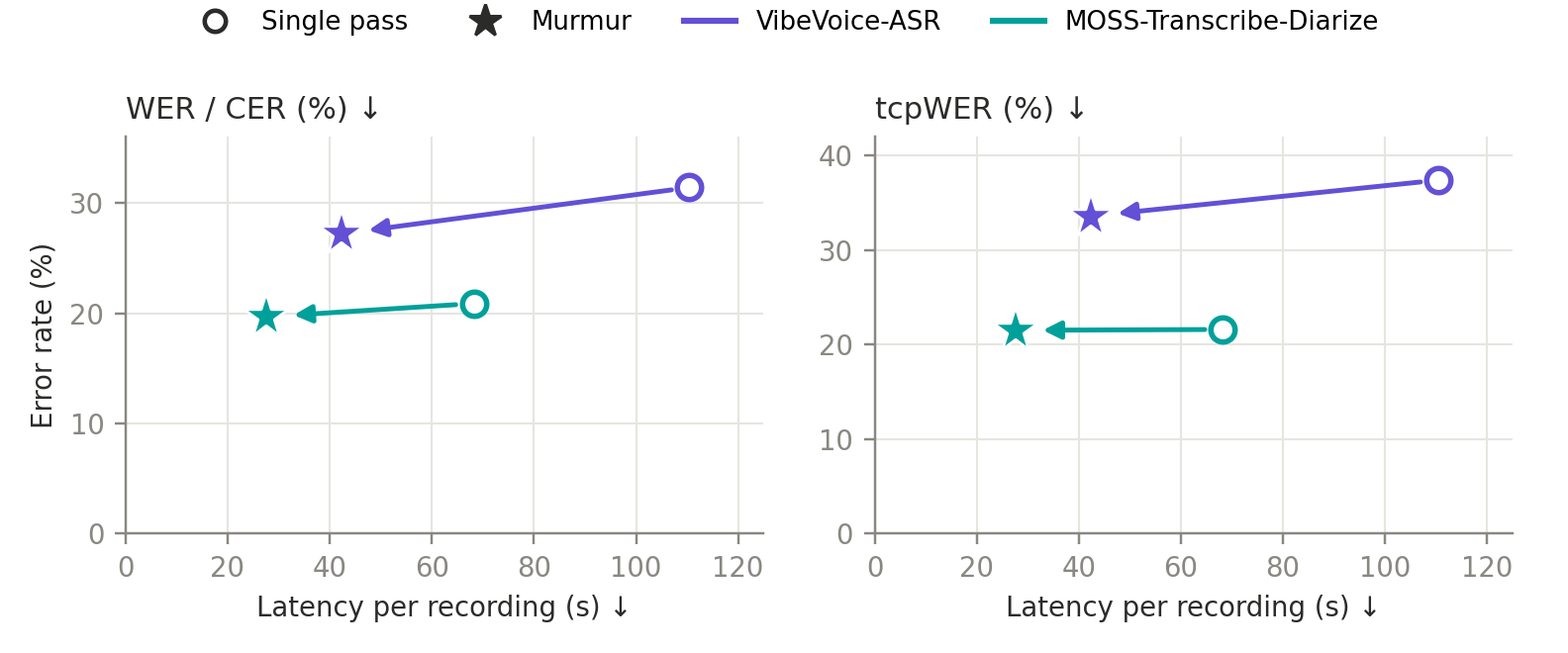}
    \caption{End-to-end latency per recording against WER/CER (left) and tcpWER (right) for each model's single pass (circles) and \name (stars) on four meeting datasets.
    Lower and further left is better (Table~\ref{tab:main}).}
    \label{fig:pareto}
\end{figure}

\begin{figure*}[t]
    \centering
    \includegraphics[width=\linewidth]{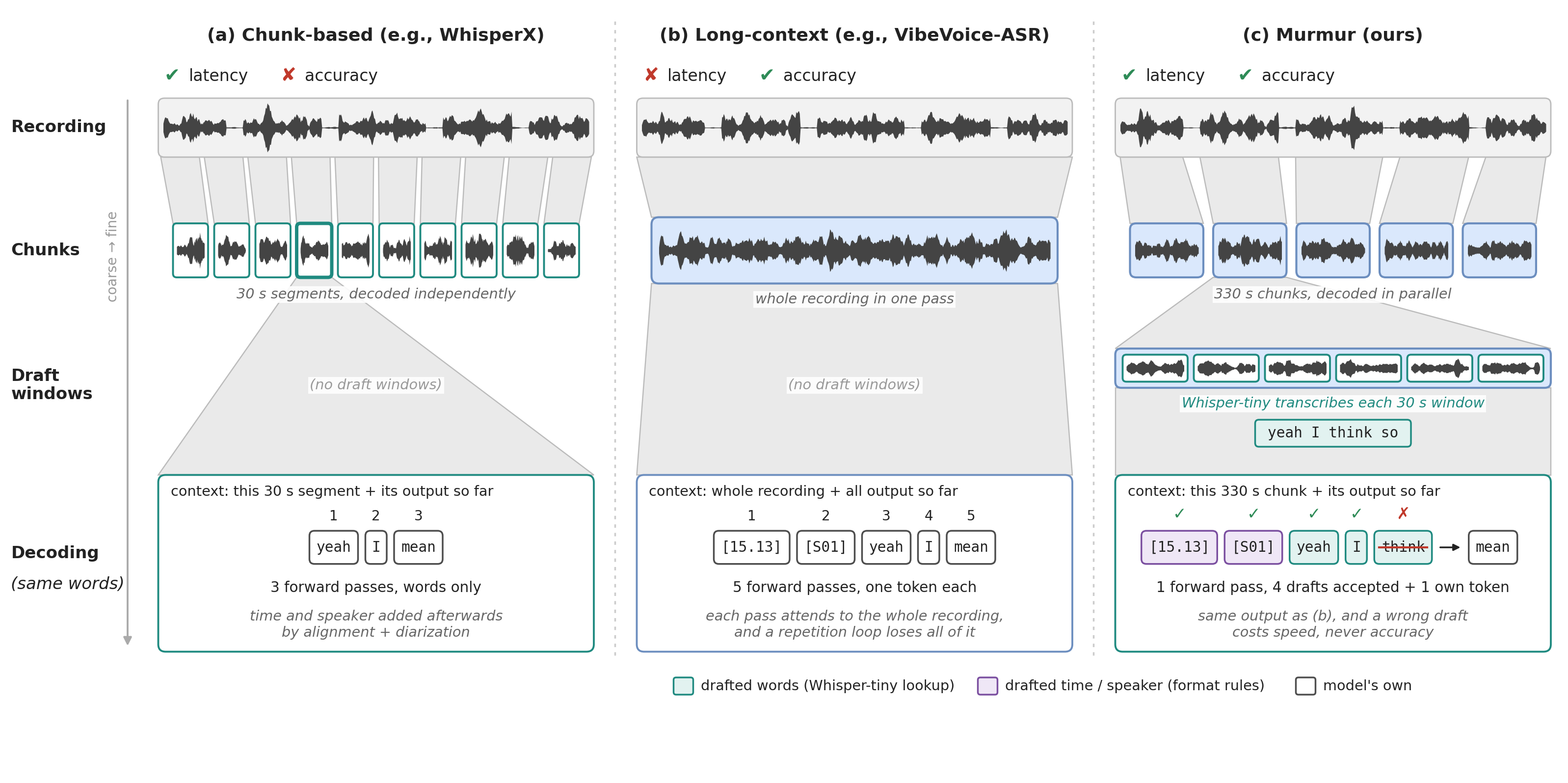}
    \caption{Long-form ASR approaches from coarse (top) to fine (bottom).
    (a) Chunk-based systems decode short segments independently and add timestamps and speakers afterwards.
    (b) Long-context models decode the whole recording in one pass, one token at a time.
    (c) \name decodes 330\,s chunks in parallel.
    Within each chunk, a short-window ASR on 30\,s windows drafts the words, format rules draft timestamps and speakers, and the long-context model verifies several drafts per step, so its output is unchanged.}
    \label{fig:headline}
\end{figure*}

Long-form automatic speech recognition (ASR) is increasingly central to many real-world speech applications, such as meetings, lectures, and clinical interactions~\cite{fox2024updated,mccowan2005ami,rousseau2012ted,chen2021gigaspeech,chiu2017speech}.

Two kinds of systems handle such audio, and neither balances accuracy and efficiency~\cite{flynn2026beyond} (\S\ref{sec:related}).
Chunk-based pipelines such as WhisperX~\cite{bain2023whisperx} transcribe 30\,s segments in parallel but lose cross-chunk context, causing errors in transcription, speaker attribution and timestamps.
Long-context models based on large language models (LLMs), such as VibeVoice-ASR and MOSS-Transcribe-Diarize~\cite{fu2026vibevoice,ai2026mosstranscribediarizetechnical}, transcribe up to an hour of audio in a single pass with speaker and timestamp tokens, but decode long sequences token by token and can fall into repetition loops.

We first treat chunk size as a tunable parameter of these long-context models (\S\ref{sec:chunksize_motivation}) and find an effective middle ground.
Chunks of 330\,s match or outperform single-pass accuracy on two models and four meeting datasets in two languages, except in speaker attribution for MOSS-Transcribe-Diarize on Mandarin, and they confine a repetition loop to one chunk (\S\ref{sec:failure}).
Yet the long-context model still decodes each chunk token by token, far more slowly than a small ASR transcribes a 30\,s window.

Based on this, we propose \name, an inference system built on \emph{hierarchical chunking} (Figure~\ref{fig:headline}).
\emph{Target chunks} of about 330\,s decide the output.
The long-context model decodes them in parallel, with enough context for accurate speaker attribution.
\emph{Draft windows} of 30\,s make decoding fast.
A small short-window ASR (e.g., Whisper-tiny) transcribes them, and its transcript serves as the draft for speculative decoding~\cite{leviathan2023fastinferencetransformersspeculative}, which the long-context model verifies (Figure~\ref{fig:system}).
Because the output interleaves words with timestamps and speaker tags, the drafter is \emph{structure-aware}.
It aligns the short-window transcript to the output through the timestamps the model writes, and drafts timestamps and speaker tags from simple format rules.
Draft windows never decide what is written, so their missing context costs speed, not accuracy.

On two models and four meeting datasets, \name cuts end-to-end latency per recording by 1.9--2.9$\times$ relative to single-pass decoding, including diarization, with output identical to greedy decoding of the same chunks in fp32 (Figure~\ref{fig:pareto}).
Most of the gain comes from parallel target chunks.
Draft windows pay off only when a decoding step is expensive, so \name enables them for VibeVoice-ASR's 7B decoder, where they cut end-to-end latency by a further 1.1--1.8$\times$, but not for MOSS-Transcribe-Diarize's 0.6B decoder (\S\ref{sec:method-when}).

Our contributions are:
\begin{enumerate}
    \item An analysis of how context length affects long-context ASR, showing that target chunks of about 330\,s preserve transcription, speaker-attribution and alignment accuracy while allowing parallel decoding.
    \item Structure-aware speculative decoding, in which a short-window ASR and format rules draft the words, timestamps, and speaker tags that the long-context model verifies.
        It is lossless and needs no training.
    \item An evaluation of \name on two long-context models and four meeting datasets in English and Mandarin.
        \name cuts end-to-end latency by 1.9--2.9$\times$ relative to single-pass decoding, and an analysis across chunk sizes, batch sizes, and decoder sizes shows when draft windows pay off.
\end{enumerate}

\section{Background}
\label{sec:related}

\subsection{Long-Form Speech Recognition}
Transformer-based ASR systems were initially trained on short utterances, with Whisper~\cite{radford2023robust} setting the de facto 30\,s context window, a limit inherited from utterance-boundary segmentation of labeled speech~\cite{radford2023robust, flynn2026beyond}.
For longer audio in meetings, lectures, and clinical settings~\cite{fox2024updated,mccowan2005ami,rousseau2012ted,chen2021gigaspeech,chiu2017speech}, chunk-based systems such as WhisperX~\cite{bain2023whisperx} split recordings into fixed 30\,s windows and stitch results with heuristics for timestamps and diarization, at the cost of boundary errors and pipeline complexity.
More recent LLM-based long-context ASR models~\cite{huo2026tagspeechendtoendmultispeakerasr,fu2026vibevoice,ai2026mosstranscribediarizetechnical} transcribe entire recordings in a single pass, natively producing speaker-attributed, timestamped output, but suffer from repetition failures and high inference latency, motivating intermediate operating points.
VibeVoice-ASR-Streaming~\cite{tu2026vibevoicestreaming} transcribes speaker-attributed speech as audio arrives but is a separately trained model.

\subsection{Efficient Speech Model Inference}
Prior work on efficient speech model inference spans model compression, optimized serving runtimes, and attention sparsity.
For the Whisper family, distillation~\cite{gandhi2023distil}, low-rank approximation~\cite{kamahori2025liteasr}, joint distillation and quantization~\cite{shao2024dq}, and KV cache architectural changes~\cite{zhang2026whispermla} reduce model size and compute, while dedicated runtimes such as faster-whisper~\cite{klein_faster_whisper} and VoxServe~\cite{kamahori2026voxserve} accelerate serving via operator fusion, quantization, dynamic batching, and streaming-aware scheduling.
Methods that exploit attention sparsity in long-context LLMs~\cite{xiao2024efficientstreaminglanguagemodels,zhang2023h2oheavyhitteroracleefficient,li2024snapkv,cai2024pyramidkv,tang2024quest,zhu2025tactic} trade accuracy for speed, whereas \name preserves accuracy.
Its speculative decoding reproduces greedy decoding (exactly in fp32), and its target chunks match or improve single-pass accuracy in most settings.

\subsection{Speculative Decoding}
A cheap drafter proposes $k$ tokens, and the target model scores all of them in one forward pass and keeps the longest prefix that matches its own greedy choice, plus one token of its own.
Under greedy decoding, the output equals the target's greedy output~\cite{leviathan2023fastinferencetransformersspeculative,chen2023acceleratinglargelanguagemodel}.
The speedup depends on how many draft tokens are accepted per forward pass and on how much a multi-token verification costs compared with a single decoding step.

A drafter need not be a model.
It can copy spans from a reference text~\cite{yang2023inference} or from the prompt~\cite{saxena2023prompt}.
\name's text drafts copy from a reference that a short-window ASR generates from the audio, aligned to the output by timestamps.
Speculative decoding has been applied to Whisper with same-family drafters~\cite{gandhi2023distil} and ASR-specific drafting schemes~\cite{wei2025specasracceleratingllmbasedautomatic,saon2026selfspeculativedecodingllmbasedasr,segalfeldman2024whispermedusasearmultihead}.
These works target transcript-only outputs.
\name instead drafts structured output (timestamps and speaker tags) for long-context models, uses a drafter with a different tokenizer, and verifies at long context.
Draft heads such as Medusa~\cite{cai2024medusa} (and its Whisper variant~\cite{segalfeldman2024whispermedusasearmultihead}) and EAGLE-3~\cite{li2025eagle3scalinginferenceacceleration} predict from the target's hidden states at every step and need training, whereas \name's drafter reads the audio once per target chunk.
Unlike multi-token prediction for LLM-based ASR~\cite{raj2024faster,lin2026paraasr}, which trains extra decoding heads into a new model, \name needs no training, keeps the original model's output exactly, drafts timestamps and speaker tags as well as words, and takes its drafts from the chunking hierarchy itself.

\section{Motivation}
\label{sec:motivation}

\begin{figure}[t]
    \centering
    \includegraphics[width=0.92\linewidth]{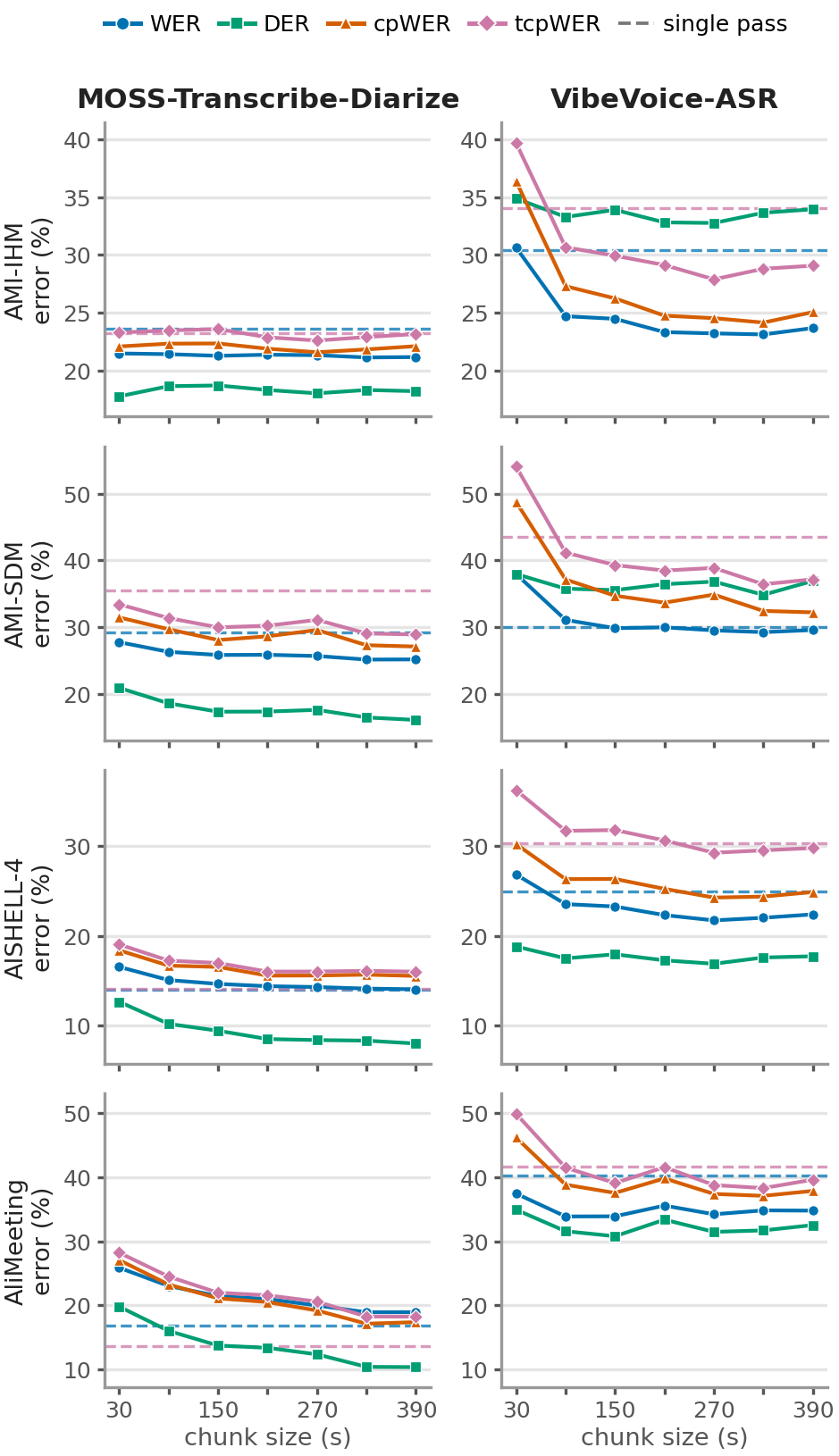}
    \caption{Error vs.\ chunk size (30--390\,s), decoded with vLLM (CER for Mandarin).
    Points below the dashed single-pass line (mean of three runs) are more accurate.
    From about 270\,s, chunking matches or beats single pass except for MOSS-Transcribe-Diarize on Mandarin, which still narrows the gap.}
    \label{fig:chunk_to_wer}
\end{figure}

\subsection{Failure Modes of Long-Form ASR}
\label{sec:failure}
Long-context ASR models can fall into a repetition loop, in which the model repeats a word or phrase and never emits an end token.
Loops can occur at any input length, even in 30\,s chunks, but a looping chunk only affects that particular chunk, whereas a looping single pass affects the whole recording.
In the batch-1 single-pass runs (the latency runs of Table~\ref{tab:main}), VibeVoice-ASR loses 1/16 AMI-IHM, 4/28 AliMeeting and 1/20 AISHELL-4 recordings this way, and MOSS-Transcribe-Diarize 2/16 AMI-IHM and 1/16 AMI-SDM recordings (Appendix~\ref{sec:appendix-failure-analysis} reports all three runs).

\subsection{How Does Longer Context Help?}
\label{sec:chunksize_motivation}
Chunk-based systems decode short windows in parallel but lose context across them.
Long-context models keep the whole recording in context, which helps most with speaker attribution~\cite{fu2026vibevoice,huo2026tagspeechendtoendmultispeakerasr}, at the cost of latency and of losing whole recordings to repetition loops (\S\ref{sec:failure}).
We therefore ask: \emph{how much context does a long-context ASR model actually need?}

Figure~\ref{fig:chunk_to_wer} varies the chunk size from 30 to 390\,s for both models on four meeting datasets, and two patterns hold across them.
First, 30\,s windows are worst for most speaker metrics (cpWER 36.4 at 30\,s vs.\ 24.1 at 330\,s for VibeVoice-ASR on AMI-IHM).
Second, accuracy plateaus by 150--330\,s, and from about 270\,s it matches or beats single pass.
For example, VibeVoice-ASR at 330\,s beats its own single pass on WER and cpWER on every dataset (AMI-SDM cpWER 32.4 vs.\ 40.9).
The one exception, speaker attribution for MOSS-Transcribe-Diarize on Mandarin, comes from linking speakers across chunks (\S\ref{sec:results}).
A few minutes of context is therefore enough, and chunks of that size can be decoded in parallel.

\subsection{Where Does the Time Go?}
\label{sec:time_motivation}
Intermediate chunks largely match single-pass accuracy, but each chunk is still decoded token by token, and decoding dominates latency.

The chunk-size analysis also points to a cheap source of drafts.
Short 30\,s windows lose speaker context, but they lose much less on the words themselves (Figure~\ref{fig:chunk_to_wer}), and a small ASR transcribes them in parallel at a fraction of the cost.
This suggests a second, finer level of chunking that proposes tokens rather than deciding them.
Two properties make speculative decoding a natural fit for it.
Decoding is memory-bound, so verifying several tokens costs about as much as generating one.
The output is also predictable, since the short-window transcript already contains most of the words, and timestamps and speaker tags follow fixed formats.
Speculative decoding also leaves the output unchanged, because it spends extra compute on verification rather than skipping computation.

\section{Method}
\label{sec:method}

\begin{figure*}[t]
    \centering
    \includegraphics[width=0.8\linewidth]{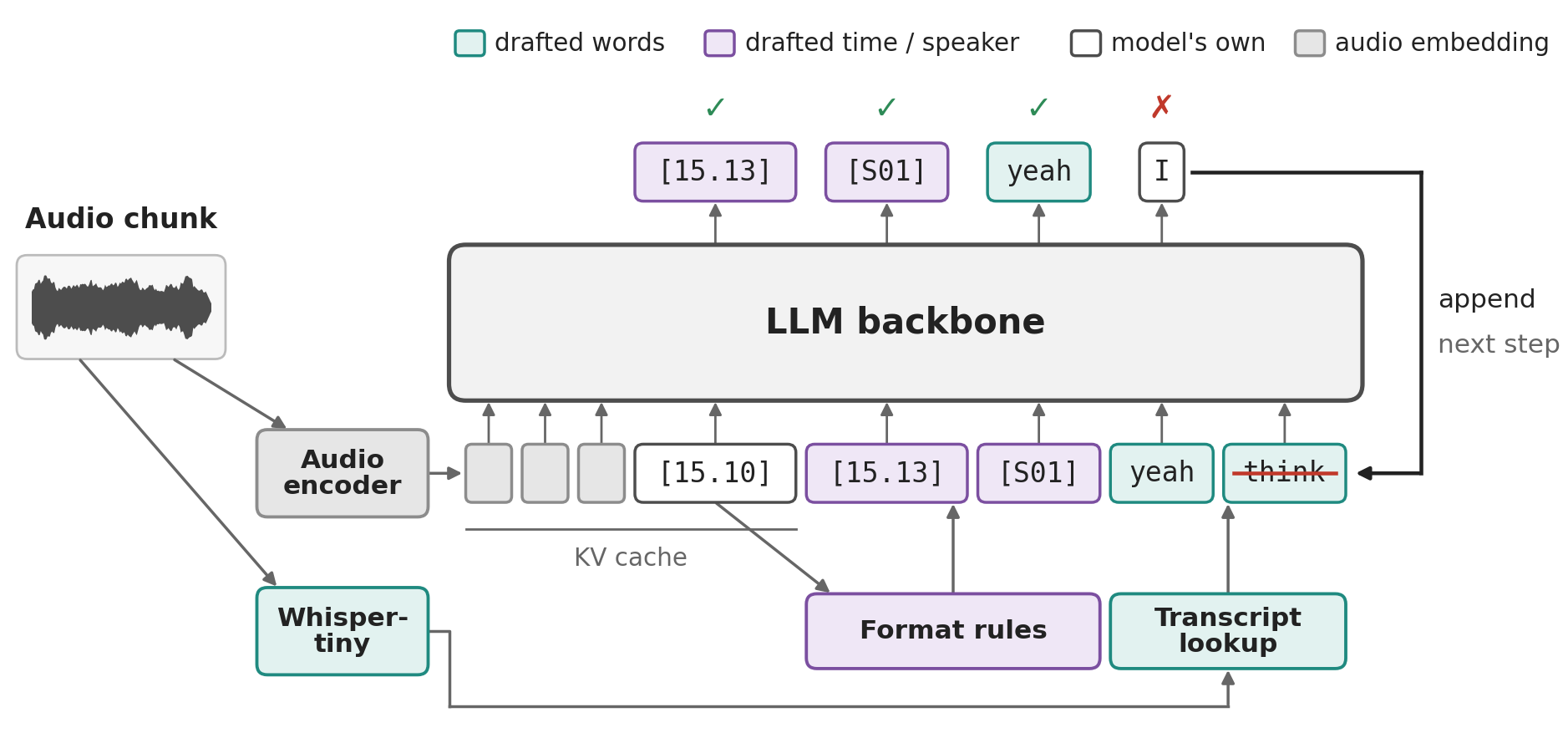}
    \caption{One decoding step of \name within a target chunk.
    The audio encoder's embeddings and the output so far are in the KV cache.
    Format rules draft the next timestamp and speaker tag from the model's own output, and a lookup in the short-window ASR (Whisper-tiny) transcript drafts the words.
    The LLM backbone scores all $k$ drafts in one forward pass.
    The matching prefix plus one token of the model's own (here \texttt{I}, replacing the rejected draft \texttt{think}) is appended to the output and starts the next step.}
    \label{fig:system}
\end{figure*}

\name chunks the recording at two granularities.
The recording is split into target chunks of up to 330\,s that the long-context model decodes in parallel (\S\ref{sec:method-chunk}).
Within each target chunk, a short-window ASR transcribes 30\,s draft windows and drafts tokens that the long-context model verifies (\S\ref{sec:method-draft}).
Chunk outputs are then stitched, with speakers linked across chunks (\S\ref{sec:method-stitch}).
Figure~\ref{fig:system} shows one decoding step within a target chunk.

\subsection{Target Chunks}
\label{sec:method-chunk}
Let $\mathcal{A}$ denote a full audio recording of duration $T$ seconds.
We first apply a voice activity detection (VAD) model to identify speech segment boundaries $\{s_1, s_2, \ldots, s_n\}$.
Segments are greedily merged to construct target chunks $\{C_1, C_2, \ldots, C_m\}$, where each chunk satisfies $|C_i| \leq c = 330$\,s, the operating point identified in \S\ref{sec:chunksize_motivation}.
A recording's chunks are decoded together as one batch, which keeps the GPU busy.

\subsection{Structure-Aware Speculative Decoding}
\label{sec:method-draft}

The drafter runs once per target chunk rather than at every decoding step, and each step's draft is a lookup.
Before each step, the drafter reads the model's output so far and determines whether the model is inside a text field or in the structure around it (Figure~\ref{fig:system}).
It then proposes up to $k=8$ tokens from one of three sources: (i) text drafts from the short-window transcript, (ii) structure drafts from format rules, and (iii) self copy from the model's own earlier output.
The model verifies every token, so drafts never decide the transcript, timestamps or speakers.

\paragraph{Text drafts.}
Whisper, run with faster-whisper, transcribes the target chunk's 30\,s draft windows with word timestamps (tiny for English, small for Mandarin).
The transcript is re-tokenized with the long-context model's tokenizer.
When the model opens a text field, the draft starts at the drafter word nearest the start time the model has just written.
Afterwards, the drafter matches the model's last 1--4 tokens in the transcript and proposes what follows.

\paragraph{Structure drafts.}
Meeting segments are usually contiguous.
For VibeVoice-ASR, after \texttt{"Start":}, the drafter proposes the previous \texttt{"End"} value.
For MOSS-Transcribe-Diarize, after a segment ends (when its closing bracket is not merged with the next one), it proposes \texttt{[$t$][$S$]}, where $t$ is the start time of the next drafter word (the previous segment's end${}+0.03$\,s if there is none) and $S$ is the model's label for the diarization speaker active at $t$ (the current speaker if unknown).

\paragraph{Self copy.}
Elsewhere, the drafter matches the model's last 2--8 tokens against its own earlier output and copies what followed (JSON keys, brackets, repeated patterns).
For VibeVoice-ASR, self copy also covers repetition loops inside a text field (\emph{loop copy}).
When the model's last tokens match its own earlier output by more tokens (at least 4) than they match the transcript, the drafter copies the model's output instead, repeated with the loop's period to fill all $k$ slots.
A loop is not in the short-window transcript, so without loop copy every transcript draft is rejected and the loop decodes one token per step.

\paragraph{Verification.}
Verification is exact greedy with KV-cache rollback.
In a batch, rows accept different numbers of tokens, which our Hugging Face engine for the fp32 checks handles by masking (Appendix~\ref{sec:appendix-impl}).

\subsection{When to Enable Draft Windows}
\label{sec:method-when}
A speculative step costs more than a plain one, because the drafter's lookup and the verification of $k{+}1$ tokens replace a single-token step (about 1.25--1.30$\times$ the cost of a plain step in vLLM).
Drafting therefore pays off only when the long-context model's decoding step is expensive.
\name always uses target chunks and enables draft windows for long-context models with a large decoder.
Both models pass audio embeddings to a decoder-only LLM and differ mainly in its size.
\name therefore enables draft windows for VibeVoice-ASR (7B) but not for MOSS-Transcribe-Diarize (0.6B), whose plain decoding of a recording's chunks already takes only 9--13\,s at batch~8, less than the diarization pass.
We still report MOSS-Transcribe-Diarize with draft windows, which confirms the rule (\S\ref{sec:results}).

\begin{table}[t]
\centering
\small
\begin{tabular}{llrrr}
\toprule
Dataset & Lang. & \#Rec. & Hours & Spk./rec. \\
\midrule
AMI-IHM & EN & 16 & 9.1 & 3.9 \\
AMI-SDM & EN & 16 & 9.1 & 3.9 \\
AliMeeting & ZH & 28 & 15.0 & 3.0 \\
AISHELL-4 & ZH & 20 & 12.7 & 5.8 \\
\bottomrule
\end{tabular}
\caption{Evaluation datasets (AMI uses the full-corpus-ASR test split).}
\label{tab:datasets}
\end{table}

\subsection{Output Stitching}
\label{sec:method-stitch}

Each chunk's timestamps are shifted by the chunk's start time in the recording.
For speakers, we keep the model's own labels \emph{within} each chunk and use diarization only to link them \emph{across} chunks.
Let $O_i(a,b)$ be the time that local speaker $a$ of chunk $C_i$ overlaps diarization speaker $b$ within $C_i$.
Local speakers are mapped to diarization speakers one-to-one by
\begin{equation}
    \pi_i = \arg\max_{\pi} \sum_{a} O_i\big(a, \pi(a)\big),
    \label{eq:linking}
\end{equation}
solved with the Hungarian algorithm.
When a chunk has more local speakers than diarization speakers, each leftover speaker $a$ joins $\arg\max_b O_i(a,b)$.
Diarization tends to under-count speakers on overlapped far-field audio, which would merge real speakers.
When it finds fewer speakers in the recording than the model reports within one chunk (counting speakers with at least 10\% of that chunk's speech), we re-run it with that count as its minimum.
No reference labels are used, and linking closes 91--100\% of the tcpWER gap between no linking and linking with reference labels (Table~\ref{tab:linking}).

\begin{table*}[t]
\centering
\small
\resizebox{\linewidth}{!}{%
\begin{tabular}{llrrrrrrrrrrrr}
\toprule
& & \multicolumn{3}{c}{AMI-IHM} & \multicolumn{3}{c}{AMI-SDM} & \multicolumn{3}{c}{AliMeeting} & \multicolumn{3}{c}{AISHELL-4} \\
\cmidrule(lr){3-5}\cmidrule(lr){6-8}\cmidrule(lr){9-11}\cmidrule(lr){12-14}
Model & System & WER & tcpWER & Lat. & WER & tcpWER & Lat. & CER & tcpWER & Lat. & CER & tcpWER & Lat. \\
\midrule
-- & WhisperX & 26.3 & 34.7 & 36.3 & 31.1 & 41.8 & 35.0 & 33.3 & 68.6 & 36.4 & 26.2 & 40.5 & 42.4 \\
\midrule
\multirow{3}{*}{VibeVoice} & Single pass & 30.4 & 34.0 & 108.9 & 30.0 & 43.6 & 99.5 & 40.3 & 41.7 & 122.5 & 25.0 & \textbf{30.3} & 110.9 \\
 & Target chunks (330\,s) & 23.8 & 29.4 & 42.2 & 29.3 & 36.5 & 40.8 & 34.6 & 39.2 & 83.6 & 23.0 & 30.8 & 68.6 \\
 & \;+ draft windows (\name) & \textbf{23.2} & \textbf{28.8} & \textbf{38.0} & \textbf{28.9} & \textbf{36.4} & \textbf{38.5} & \textbf{34.1} & \textbf{38.7} & \textbf{47.1} & \textbf{22.6} & 30.5 & \textbf{45.6} \\
\midrule
\multirow{3}{*}{MOSS} & Single pass & 23.6 & 23.2 & 54.2 & 29.2 & 35.5 & 77.1 & \textbf{16.8} & \textbf{13.6} & 90.8 & \textbf{14.0} & \textbf{14.1} & 50.8 \\
 & Target chunks (\name) & \textbf{21.6} & \textbf{23.0} & \textbf{26.4} & \textbf{25.3} & \textbf{29.0} & \textbf{26.4} & 17.4 & 17.5 & \textbf{31.6} & 14.3 & 16.6 & \textbf{26.1} \\
 & \textcolor{gray}{\;+ draft windows} & \textcolor{gray}{21.6} & \textcolor{gray}{23.0} & \textcolor{gray}{30.2} & \textcolor{gray}{25.3} & \textcolor{gray}{29.0} & \textcolor{gray}{31.4} & \textcolor{gray}{17.4} & \textcolor{gray}{17.6} & \textcolor{gray}{38.8} & \textcolor{gray}{14.3} & \textcolor{gray}{16.6} & \textcolor{gray}{35.3} \\
\bottomrule
\end{tabular}}
\caption{Accuracy (\%) and end-to-end latency (Lat., mean seconds per recording, including diarization for target chunks).
\name is target chunks plus draft windows for VibeVoice-ASR and target chunks alone for MOSS-Transcribe-Diarize (\S\ref{sec:method-when}).
The gray row shows MOSS-Transcribe-Diarize with draft windows for reference.
Single-pass accuracy is the mean of three runs.
On AliMeeting, tcpWER can fall below CER, which is scored on one time-ordered stream rather than per speaker.}
\label{tab:main}
\end{table*}

\section{Experimental Setup}
\label{sec:eval}

\subsection{Models and System}
We evaluate VibeVoice-ASR ($\sim$8.5B parameters, Qwen2.5-7B decoder) and MOSS-Transcribe-Diarize ($\sim$0.9B, with a 24-layer audio encoder and a 0.6B decoder).
The drafter is Whisper-tiny ($\sim$39M) for English and Whisper-small ($\sim$244M) for Mandarin, run with faster-whisper~\cite{klein_faster_whisper} on the same GPU as the long-context model.
Voice activity detection and diarization use pyannote~\cite{bredin2019pyannoteaudioneuralbuildingblocks}.
All experiments run on one NVIDIA H100 80\,GB GPU in bf16 with greedy decoding and $k=8$ draft tokens, on vLLM (batch 1--16) with our drafter implemented as a vLLM speculative-decoding proposer.
All experiments in this paper took about 200 GPU-hours in total.

In bf16, batched kernels are not bitwise deterministic, so near-ties occasionally flip a token or whether a chunk enters a repetition loop.
\name's differences from greedy decoding stay within greedy decoding's own spread across batch sizes and repeated runs, apart from one chunk that entered a loop (Table~\ref{tab:noise}), and in fp32 its output is identical (Appendix~\ref{sec:appendix-results}).
We report single runs, except that single-pass accuracy is the mean of three runs (batch 1, 4 and 8), because whether a single pass enters a loop varies between runs.

\subsection{Datasets}
We use four meeting datasets in English and Mandarin (Table~\ref{tab:datasets}), namely AMI~\cite{mccowan2005ami} with headset (IHM) and distant (SDM) microphones, AliMeeting~\cite{yu2022m2met} and AISHELL-4~\cite{fu2021aishell4}.

\subsection{Metrics}
We report four error rates (\%, lower is better):
\begin{itemize}[nosep,leftmargin=1.2em]
    \item \textbf{WER} (CER for Mandarin): the words alone, ignoring speakers.
    \item \textbf{DER}: the share of audio attributed to the wrong speaker.
    \item \textbf{cpWER}~\cite{watanabe2020chime}: WER per speaker, after the best mapping of output to reference speakers.
    \item \textbf{tcpWER}~\cite{vonneumann2023meeteval}: cpWER that also requires each word to fall within 5\,s of its reference time.
        It is our primary metric, since it captures transcription, speaker attribution and alignment together.
\end{itemize}
Each metric is computed per recording and averaged over a dataset's recordings.
We compute cpWER and tcpWER with MeetEval, WER with jiwer, and DER with pyannote.metrics~\cite{bredin2017pyannotemetrics} (0.25\,s collar, overlap scored), with formulas in Appendix~\ref{sec:appendix-metrics}.
A recording or chunk fails when it enters a repetition loop and never closes its output.
We score whatever was decoded, without retries, so a failed VibeVoice-ASR output (unparseable JSON) scores 100\%, while MOSS-Transcribe-Diarize keeps the words decoded before the loop.

For speed, \textbf{decoding time} is wall-clock time for a whole dataset at a fixed batch size, with speculative decoding compared against greedy decoding on the same engine, batch and chunks (\name's time includes the drafter).
\textbf{End-to-end latency} is measured per recording.
Recordings are processed one at a time, and a recording's target chunks (about 7) are decoded in parallel at batch~8.
For target chunks it includes the pyannote VAD and diarization pass (14.9--17.2\,s, added serially, which is conservative) and speaker linking's re-diarization (0--3.9\,s).
Single pass runs at batch~1 and needs no diarization, and WhisperX's time includes its own alignment and diarization.

\subsection{Baselines}
We compare against single pass, WhisperX (30\,s windows with Whisper large-v3 and pyannote), chunked greedy decoding with the same chunks, and vLLM's built-in n-gram speculative decoding (prompt lookup decoding~\cite{saxena2023prompt}, $k=8$, 2--4-token matches).

\section{Results}
\label{sec:results}

\subsection{Accuracy}
At 330\,s, chunked decoding matches or beats single pass on both models and all four datasets, with one exception (Table~\ref{tab:main}).
For MOSS-Transcribe-Diarize on Mandarin, CER stays close to single pass but tcpWER is 2.5--3.9 points worse, mostly from linking speakers across chunks (Table~\ref{tab:linking}).
\name beats WhisperX on tcpWER on every dataset.
Draft windows keep this accuracy, since speculative decoding is lossless.
Their small differences from target chunks alone in Table~\ref{tab:main} are bf16 nondeterminism, and in fp32 the outputs are identical (Table~\ref{tab:lossless}).

\subsection{Repetition Loops}
Chunking confines repetition loops.
Over three single-pass runs, VibeVoice-ASR loses whole recordings to loops on every dataset (2--19\% of runs) and MOSS-Transcribe-Diarize on both AMI datasets.
With target chunks, a loop costs one chunk instead (Appendix~\ref{sec:appendix-failure-analysis}).
This is where chunking gains the most accuracy.
Draft windows also speed through loops, because the model's own recent output, which self copy proposes, is exactly what a loop repeats.
This is why speculative decoding gains most on Mandarin, where loops are most common.

\subsection{Latency}
Decoding target chunks in parallel gives most of the speedup.
On AMI-IHM, VibeVoice-ASR drops from 108.9 to 42.2\,s per recording, and draft windows cut it further to 38.0\,s.
For MOSS-Transcribe-Diarize, target chunks alone are fastest, and draft windows would make it 14--35\% slower, as \S\ref{sec:method-when} predicts.

\subsection{Where the Speedup Comes From}
The three draft sources add up.
For VibeVoice-ASR on AMI-IHM at batch~8 (decoding time, Table~\ref{tab:vllm}), text drafts alone give 1.40$\times$, structure drafts raise this to 1.58$\times$, and self copy to 1.83$\times$.
At batch~1, \name is faster than vLLM's n-gram decoding on every dataset for both models (Table~\ref{tab:ngram_compare}).
N-gram decoding pulls ahead only on Mandarin at larger batches (Table~\ref{tab:vllm}), because it copies the model's own output, which is what a repetition loop repeats, and loops are most common there.

Drafting pays off when a plain decoding step is expensive relative to the overhead of a speculative step (\S\ref{sec:method-when}).
Longer chunks make each verification costlier, yet the speedup holds from 150 to 510\,s and peaks at 330\,s on AMI-IHM (Table~\ref{tab:chunksize}).
Decoder size matters more.
\name speeds up VibeVoice-ASR's 7B decoder at every batch size, but not MOSS-Transcribe-Diarize's 0.6B decoder, whose steps are already cheap (\S\ref{sec:method-when}).

\begin{table}[t]
\centering
\small
\resizebox{\linewidth}{!}{%
\begin{tabular}{llrrrr}
\toprule
Model & Method & IHM & SDM & Ali & AISH \\
\midrule
\multirow{2}{*}{VibeVoice} & n-gram & 1.37 & 1.22 & 1.85$^\dagger$ & 1.34 \\
 & \name & \textbf{1.78} & \textbf{1.47} & \textbf{2.03}$^\dagger$ & \textbf{1.43} \\
\midrule
\multirow{2}{*}{MOSS} & n-gram & 0.69 & 0.68 & 0.74 & 0.67 \\
 & \name & \textbf{0.91} & \textbf{0.82} & \textbf{0.85} & \textbf{0.79} \\
\bottomrule
\end{tabular}}
\caption{Speedup in decoding time over greedy decoding for vLLM n-gram and \name at batch~1, with 330\,s chunks over the whole dataset.
At batch~1, chunks are decoded one at a time, so each speedup is the per-chunk gain.
Times include \name's drafter, which runs alongside decoding (Appendix~\ref{sec:appendix-impl}), and exclude the diarization pass, which is the same for all.
Batch~4, 8 and 16 are in Table~\ref{tab:vllm}.
$^\dagger$Greedy baseline inflated by repetition loops (13 of 177 chunks).}
\label{tab:ngram_compare}
\end{table}

\section{Discussion}
Two practical guidelines follow, one per level of chunking.
For accuracy, size target chunks at a few minutes rather than at either extreme, since 30\,s windows lose speaker context and a single pass risks losing whole recordings to repetition loops.
Applications that need only the words can use shorter target chunks, since WER saturates earlier than the speaker metrics.

For speed, drafting from short windows is worth enabling when the long-context model's decoding step is expensive, as for VibeVoice-ASR, and its output is predictable.
For a small decoder, each step is already cheap, and the overhead of a speculative step cancels the gain.
Loop copy speeds through repetition loops but does not prevent them.
Preventing loops, and better cross-chunk speaker linking for Mandarin, would improve both accuracy and speed.

\section{Conclusion}
\label{sec:conclusion}
We asked how much context long-form ASR models need and found that 330\,s chunks match or beat single-pass accuracy on two models and four meeting datasets, except in speaker attribution for MOSS-Transcribe-Diarize on Mandarin, while avoiding single-pass failures.
Building on this, \name uses hierarchical chunking.
The long-context model decodes 330\,s target chunks in parallel, and a short-window ASR on 30\,s draft windows drafts the words while format rules draft the timestamps and speaker tags.
The model verifies every token, so its output is unchanged.
\name has 1.9--2.9$\times$ lower end-to-end latency per recording than a single pass, including diarization, and is more accurate on AMI.
Most of this gain comes from the target chunks, and draft windows add 1.1--1.8$\times$ for VibeVoice-ASR.

\section*{Limitations}
\name is designed for long-context ASR models that natively produce timestamped, speaker-attributed transcriptions, and we evaluate two such models.
Gains from the draft windows depend on drafter quality (lower on Mandarin audio).
Draft windows also cannot prevent repetition loops, and they require structure rules per output format.
For a small decoder in an optimized engine (MOSS-Transcribe-Diarize in vLLM), they turn into a slowdown, so \name enables them per model (\S\ref{sec:method-when}).
Speaker linking re-runs diarization with a speaker-count hint whose 10\% speech-share threshold we chose on the evaluation meetings.
MOSS-Transcribe-Diarize lists AISHELL-4 among its training sources, which may favor its single pass there.
We report single runs, and single-pass VibeVoice-ASR in particular varies between runs.

\section*{Ethics Statement}
\paragraph{Licenses and intended use.}
All artifacts used in this work are publicly released under licenses that permit research use, and our usage is consistent with their intended uses.
For datasets, we use the AMI Meeting Corpus (CC BY 4.0)~\cite{mccowan2005ami}, AliMeeting (CC BY-SA 4.0), and AISHELL-4 (CC BY-SA 4.0).
For models and software, we use VibeVoice-ASR (MIT License)~\cite{fu2026vibevoice}, MOSS-Transcribe-Diarize (Apache 2.0)~\cite{ai2026mosstranscribediarizetechnical}, Whisper (MIT)~\cite{radford2023robust}, faster-whisper, WhisperX (BSD-4-Clause)~\cite{bain2023whisperx}, pyannote.audio (MIT), and vLLM (Apache 2.0)~\cite{kwon2023efficient}.
We release our \name implementation under the Apache 2.0 license.

\paragraph{Data and risks.}
All experiments involve transcription of pre-existing, publicly available recordings and do not involve human subjects or the collection of new data.
Automatic transcription systems of the kind studied here could be misused for surveillance.
We encourage responsible deployment with appropriate consent and access controls.

\bibliographystyle{acl_natbib}
\bibliography{custom}

\appendix

\section{Baseline Configuration}
\label{sec:appendix-baseline}
\subsection{WhisperX}
We compare against WhisperX~\cite{bain2023whisperx} using Whisper large-v3 as the backbone, running in \texttt{float16} precision on the same GPU as \name.
The original WhisperX paper reports results with Whisper large-v2, but we instead use large-v3 because it is a stronger and more recent checkpoint, is the current default in the upstream WhisperX codebase, and gives WhisperX its best published WER on long-form English meeting audio, making it the most competitive baseline for our comparison.

The pipeline follows the four-stage structure documented in the official WhisperX repository: (i) batched ASR with the \texttt{faster-whisper} backend at an internal batch size of 16 and the language set to English or Mandarin, (ii) forced phoneme-level alignment with a \texttt{wav2vec2} model to recover word-level timestamps, (iii) speaker diarization with \texttt{pyannote.audio} via the \texttt{DiarizationPipeline}, and (iv) word-to-speaker assignment using the aligned timestamps.

\subsection{vLLM N-gram Speculative Decoding}
We use vLLM's prompt-lookup proposer with $k=8$ and 2--4-token matches, with the same server settings as greedy decoding.
Enabling speculative decoding disables vLLM's asynchronous scheduling.
VibeVoice-ASR is served with vLLM 0.14.0 and MOSS-Transcribe-Diarize with a vLLM 0.23.1 nightly build (\texttt{0.23.1rc1.dev949}), the version its model code requires.
Greedy, n-gram and \name use the same server build for each model.

\section{Implementation Details}
\label{sec:appendix-impl}

\subsection{Drafter}
The drafter is Whisper run with faster-whisper~\cite{klein_faster_whisper} (1.2.1, CTranslate2 4.7.2) in float16 on the same GPU as the long-context model, using its batched pipeline (batch 16 over 30\,s windows), greedy decoding (beam 1) and word timestamps.
English uses Whisper-tiny.
Mandarin uses Whisper-small with \texttt{language="zh"}, a short prompt in punctuated simplified Chinese (both long-context models write simplified characters and Chinese punctuation, so unpunctuated drafts are rejected at every comma), and traditional-to-simplified conversion of the output.
The drafter transcript is re-tokenized with the long-context model's tokenizer.
English is re-tokenized word by word, so the token offset of each word is exact.
Mandarin is re-tokenized as one string, since the model's tokens merge across the drafter's 1--2-character words.
For VibeVoice-ASR the text is JSON-escaped, since it appears inside a JSON string.
An index of all 1--4-grams over the drafter tokens supports lookup.

\subsection{Draft Rules}
Before each verification step, the drafter reads the model's output so far and tracks whether it is inside a text field.
Table~\ref{tab:draft_rules} lists the rule used in each state.
The first rule that produces tokens is used, and at most $k=8$ tokens are proposed.

\begin{table*}[t]
\centering
\small
\begin{tabular}{p{0.2\linewidth}p{0.36\linewidth}p{0.36\linewidth}}
\toprule
State & VibeVoice-ASR (JSON) & MOSS-Transcribe-Diarize (\texttt{[start][Sxx] text[end]}) \\
\midrule
Text field opens & If the model's last 2--8 tokens occur earlier in its own output, copy what followed them (this catches repeated segments). Otherwise, drafter words from the word nearest the segment's \texttt{"Start"} & Same, using the segment's \texttt{[start]} \\
First word of the text & Find that word among the drafter words within $\pm$8 words of the start time and continue from there (English only) & Same \\
Rest of the text & Longest 1--4-gram match of the model's last tokens in the drafter transcript, continuing from the match nearest the previous position. If the model's own output matches more tokens ($\geq$4), as in a repetition loop, copy it instead, repeated with the loop's period (loop copy) & Same, without loop copy \\
Segment ends & After \texttt{"Start":}, propose the previous \texttt{"End"} value and \texttt{,"End":} & After \texttt{[end]} (when its \texttt{]} is not merged with the next \texttt{[}), propose \texttt{[t][S]}, where $t$ is the start time of the next drafter word (fallback: end $+\,0.03$\,s) and $S$ is the model's label for the diarization speaker active at $t$ (fallback: same speaker) \\
Elsewhere & Copy what followed the longest 2--8-token match of the model's last tokens in its own earlier output & Same \\
\bottomrule
\end{tabular}
\caption{Draft rules.
All drafts are verified, so a wrong rule costs speed, never accuracy.
MOSS-Transcribe-Diarize's tokenizer merges \texttt{][} and \texttt{.[} into single tokens, so its rules match the decoded text rather than token boundaries.
The Hugging Face experiments use a simpler MOSS-Transcribe-Diarize rule (previous end $+\,0.03$\,s, same speaker) and 2--4-token self copy.}
\label{tab:draft_rules}
\end{table*}

\subsection{Serving with vLLM}
\paragraph{Proposer.}
For MOSS-Transcribe-Diarize, the drafter is a custom vLLM proposer class.
vLLM 0.14.0 has no custom-proposer hook, so for VibeVoice-ASR we start the server with the n-gram method and replace the n-gram proposer's methods with ours at import time.
vLLM does not pass request IDs to a proposer, so we wrap the model runner's draft step to hand over the batch's request IDs.
The client tags each chunk's request with an \texttt{X-Request-Id} header and writes that chunk's drafter transcript to a file named by the tag, and the proposer loads it when the request first appears.
A request without a transcript still receives structure and self-copy drafts.

\paragraph{Pipelining.}
All chunks of a dataset are submitted through one client thread pool whose size equals the batch size (the server's \texttt{max-num-seqs}), so the server holds a steady batch.
A drafter thread transcribes the chunks in submission order, and each chunk's request is sent as soon as its transcript is ready, so drafting overlaps with decoding of earlier chunks.
Reported time is the wall time for the whole dataset, including the drafter, and one untimed warm-up request precedes it.

\paragraph{Server settings.}
VibeVoice-ASR uses bf16, maximum model length 65{,}536, 32{,}768 batched tokens, chunked prefill, no prefix caching, CUDA graphs, and GPU memory utilization 0.85 (0.80 when the drafter shares the GPU).
MOSS-Transcribe-Diarize uses 49{,}152 batched tokens (131{,}072 with chunked prefill for single pass) and memory utilization 0.85.
Decoding is greedy (temperature 0).
The output-token budget for MOSS-Transcribe-Diarize is $\max(512,\,32\,\text{tokens/s} \times \text{duration})$.
For VibeVoice-ASR, it is $\min(32{,}768,\,\max(256,\,100\,\text{tokens/s} \times \text{duration}))$.
A chunk that reaches the budget without closing its output is flagged as failed.

\subsection{Hugging Face Engine}
The fp32 losslessness checks use our own Hugging Face decoding loop with exact greedy verification and KV-cache rollback.
\paragraph{Batched verification.} Rows of a batch accept different numbers of tokens.
Rejected slots are masked rather than cropped, positions are counted over unmasked tokens only, and the cache is compacted every 32 steps.
\paragraph{Split attention.} Above 8{,}192 cached tokens, each multi-token verification is split into (a) attention of the new tokens to the past cache, without a mask, using flash attention on queries regrouped per KV head (no \texttt{repeat\_kv}), which also returns the log-sum-exp; (b) the causal block among the $\leq k{+}1$ new tokens, computed directly; and (c) an exact log-sum-exp merge of (a) and (b).
Against reference attention the fp32 error is at most $2\times10^{-7}$.
Without it, a masked call cannot use flash attention, and the memory-efficient fallback, which parallelizes only over query tiles, leaves the GPU mostly idle on a 42k-token cache (56.6 of 77.8\,ms per verification).
\paragraph{Determinism.} cuDNN attention is disabled in all arms, because it re-plans for every new verification shape (about 27\,ms per call).
VibeVoice-ASR's acoustic tokenizer samples its latents, so we seed the random generator before each chunk's prefill, since otherwise two arms diverge from the first token.

\section{Additional Results}
\label{sec:appendix-results}

\begin{table}[t]
\centering
\small
\resizebox{\linewidth}{!}{%
\begin{tabular}{lrrrr}
\toprule
\textbf{Method} & \textbf{WER} & \textbf{DER} & \textbf{tcpWER}& \textbf{Latency} \\
\midrule
WhisperX        & 26.3\%& 27.4\%& 34.7\%& \textbf{36.3\,s}\\
VibeVoice-ASR   & 27.4\%& 42.2\%& 31.6\%& 108.9\,s\\
MOSS            & \textbf{23.7\%}& \textbf{19.2\%}& \textbf{23.3\%}& 54.2\,s\\
\bottomrule
\end{tabular}}
\caption{Single-pass error rates and latency per recording on AMI-IHM~\cite{mccowan2005ami}, from the batch-1 runs (the latency runs of Table~\ref{tab:main}, whose single-pass accuracy is the mean of three runs).
A recording that enters a repetition loop keeps what it decoded (VibeVoice-ASR's JSON then fails to parse and scores 100\% WER).
VibeVoice-ASR loses 1/16 recordings and MOSS-Transcribe-Diarize 2/16 to loops.
Without them, WER, DER and tcpWER are 22.6\%, 38.3\% and 27.0\% for VibeVoice-ASR and 19.8\%, 14.8\% and 19.3\% for MOSS-Transcribe-Diarize.}
\label{tab:baseline_comparison}
\end{table}

\paragraph{Losslessness.}
In fp32, \name's output is identical to greedy decoding (text, timestamps and speaker labels) for VibeVoice-ASR on AMI-IHM EN2002a 120--420\,s and on ES2004b 120--720\,s, a 10-minute chunk whose cache exceeds the 8{,}192-token split-attention threshold, and for MOSS-Transcribe-Diarize on EN2002a 120--420\,s.
The bf16 differences in the main tables come from batched kernels that are not bitwise deterministic.
Table~\ref{tab:lossless} compares \name with greedy decoding on the same engine (vLLM, batch~8, 330\,s chunks), and Table~\ref{tab:noise} puts the differences next to greedy decoding's own spread across batch sizes and repeated runs.

\begin{table}[t]
\centering
\small
\resizebox{\linewidth}{!}{%
\begin{tabular}{llrrrr}
\toprule
Model & Decoding & IHM & SDM & Ali & AISH \\
\midrule
\multirow{2}{*}{VibeVoice} & Greedy & 23.8 / 29.4 & 29.3 / 36.5 & 34.6 / 39.2 & 23.0 / 30.8 \\
 & \name & 23.2 / 28.8 & 28.9 / 36.4 & 34.1 / 38.7 & 22.6 / 30.5 \\
\midrule
\multirow{2}{*}{MOSS} & Greedy & 21.6 / 23.0 & 25.3 / 29.0 & 17.4 / 17.5 & 14.3 / 16.6 \\
 & \name & 21.6 / 23.0 & 25.3 / 29.0 & 17.4 / 17.6 & 14.3 / 16.6 \\
\bottomrule
\end{tabular}}
\caption{\name vs.\ greedy decoding on the same engine, as WER (CER for Mandarin) / tcpWER (\%) with vLLM, batch~8 and 330\,s chunks.
Differences are bf16 batch nondeterminism, and in fp32 the outputs are identical.}
\label{tab:lossless}
\end{table}

\begin{table*}[t]
\centering
\small
\begin{tabular}{llccccc}
\toprule
& & \multicolumn{2}{c}{Greedy spread} & \multicolumn{3}{c}{\name $-$ greedy} \\
\cmidrule(lr){3-4}\cmidrule(lr){5-7}
Model & Data & batch 1/4/8 & repeats & batch 1 & batch 4 & batch 8 \\
\midrule
\multirow{4}{*}{VibeVoice} & IHM & 0.68 / 1.38 & 0.71 / 0.76 & $-$0.70 / $-$1.14 & +0.02 / +0.45 & $-$0.68 / $-$0.67 \\
 & SDM & 0.36 / 0.74 & 0.14 / 0.66 & +0.30 / +0.61 & +0.98$^\dagger$ / +0.26 & $-$0.36 / $-$0.09 \\
 & Ali & 1.13 / 1.06 & 1.82 / 1.40 & +0.11 / +0.08 & +0.13 / $-$0.05 & $-$0.55 / $-$0.50 \\
 & AISH & 2.42 / 2.34 & 1.35 / 1.47 & +0.53 / +0.47 & +0.85 / +0.52 & $-$0.34 / $-$0.27 \\
\midrule
\multirow{4}{*}{MOSS} & IHM & 0.75 / 0.94 & 0.02 / 0.12 & 0.00 / +0.01 & +0.03 / 0.00 & $-$0.01 / $-$0.01 \\
 & SDM & 0.22 / 0.23 & 0.02 / 0.18 & 0.00 / +0.02 & $-$0.03 / $-$0.02 & $-$0.04 / +0.06 \\
 & Ali & 0.02 / 0.03 & 0.03 / 0.03 & 0.00 / 0.00 & +0.01 / +0.04 & +0.01 / +0.04 \\
 & AISH & 0.02 / 0.25 & 0.03 / 0.02 & 0.00 / +0.02 & 0.00 / $-$0.11 & $-$0.02 / $-$0.07 \\
\bottomrule
\end{tabular}
\caption{bf16 noise floor, WER (CER for Mandarin) / tcpWER points.
Greedy spread: range of greedy decoding over batch 1, 4 and 8, and over three runs at batch~8.
\name $-$ greedy: difference at the same batch size (330\,s chunks, vLLM).
$^\dagger$One chunk entered a repetition loop (EN2002b).}
\label{tab:noise}
\end{table*}

\paragraph{Speaker linking.}
Table~\ref{tab:linking} compares cross-chunk speaker linking on the same \name outputs (330\,s, vLLM, batch~8) under no linking (each chunk's local labels reused across chunks), an independent best-overlap rule, our linking (\S\ref{sec:method-stitch}), and linking with the reference speaker labels.
WER is identical in every column, so only speaker attribution differs.
Our linking closes 91--100\% of the gap between no linking and reference linking.
The remaining gap is largest on AliMeeting, where diarization confuses speaker identities in a few heavily overlapped recordings.
With reference linking, chunked MOSS-Transcribe-Diarize is still about 1 tcpWER point behind its single pass on Mandarin, which is the cost of decoding each chunk without the rest of the recording.

\begin{table}[t]
\centering
\small
\resizebox{\linewidth}{!}{%
\begin{tabular}{llrrrrr}
\toprule
Model & Data & None & Overlap & Ours & Ref. & Closed \\
\midrule
\multirow{4}{*}{VibeVoice} & IHM & 97.9 & 29.5 & 28.8 & 28.6 & 99.7\% \\
 & SDM & 102.7 & 40.0 & 36.4 & 36.9 & 100.8\% \\
 & Ali & 76.4 & 40.0 & 38.7 & 35.1 & 91.4\% \\
 & AISH & 107.4 & 30.6 & 30.5 & 29.9 & 99.2\% \\
\midrule
\multirow{4}{*}{MOSS} & IHM & 98.0 & 23.5 & 23.0 & 22.8 & 99.7\% \\
 & SDM & 99.9 & 30.6 & 29.0 & 28.1 & 98.7\% \\
 & Ali & 68.7 & 23.7 & 17.6 & 14.6 & 94.6\% \\
 & AISH & 105.3 & 17.7 & 16.6 & 15.3 & 98.6\% \\
\bottomrule
\end{tabular}}
\caption{Cross-chunk speaker linking, tcpWER (\%) on the same \name outputs.
None: local labels reused across chunks.
Overlap: each local speaker independently takes its best-overlapping diarization speaker.
Ref.: linking with the reference speaker labels (by overlap, so not the tcpWER optimum, and ours is lower on VibeVoice AMI-SDM).
Closed: share of the None$\to$Ref.\ gap that ours closes.}
\label{tab:linking}
\end{table}

\paragraph{Chunk size.}
Table~\ref{tab:chunksize} reports the speedup over greedy decoding at each chunk size with vLLM at batch~1 on all 16 AMI meetings, and Table~\ref{tab:chunksize_acc} the corresponding accuracy.
Accuracy is unchanged at every size up to bf16 noise.
For VibeVoice-ASR the speedup peaks at 330\,s on AMI-IHM and changes little on AMI-SDM, while for MOSS-Transcribe-Diarize it grows with chunk size but stays near or below 1.

\begin{table}[t]
\centering
\small
\begin{tabular}{llrrr}
\toprule
Model & Data & 150\,s & 330\,s & 510\,s \\
\midrule
VibeVoice & IHM & 1.58$\times$ & \textbf{1.78$\times$} & 1.58$\times$ \\
VibeVoice & SDM & 1.43$\times$ & 1.47$\times$ & \textbf{1.51$\times$} \\
MOSS & IHM & 0.87$\times$ & 0.91$\times$ & \textbf{1.03$\times$} \\
MOSS & SDM & 0.81$\times$ & 0.82$\times$ & \textbf{0.94$\times$} \\
\bottomrule
\end{tabular}
\caption{\name speedup over greedy decoding at the same chunk size (vLLM, batch~1, whole dataset including the drafter, all 16 AMI meetings).}
\label{tab:chunksize}
\end{table}

\begin{table*}[t]
\centering
\small
\setlength{\tabcolsep}{5pt}
\begin{tabular}{llccc}
\toprule
Model & Data & 150\,s & 330\,s & 510\,s \\
\midrule
VibeVoice & IHM & 25.55 $\to$ 25.69 / 28.83 $\to$ 29.25 & 24.90 $\to$ 24.30 / 29.50 $\to$ 28.36 & 24.14 $\to$ 23.98 / 27.94 $\to$ 27.99 \\
VibeVoice & SDM & 34.83 $\to$ 34.62 / 39.90 $\to$ 39.67 & 32.59 $\to$ 32.98 / 37.03 $\to$ 37.64 & 32.14 $\to$ 32.01 / 37.11 $\to$ 36.97 \\
MOSS & IHM & 23.62 $\to$ 23.60 / 24.83 $\to$ 24.81 & 21.87 $\to$ 21.88 / 22.96 $\to$ 22.97 & 22.12 $\to$ 22.14 / 23.14 $\to$ 23.15 \\
MOSS & SDM & 30.12 $\to$ 30.13 / 31.98 $\to$ 31.98 & 27.23 $\to$ 27.25 / 28.73 $\to$ 28.74 & 27.27 $\to$ 27.19 / 28.80 $\to$ 28.73 \\
\bottomrule
\end{tabular}
\caption{Accuracy of greedy decoding $\to$ \name for the runs in Table~\ref{tab:chunksize}, as cpWER / tcpWER (\%).
Differences are bf16 nondeterminism.}
\label{tab:chunksize_acc}
\end{table*}

\paragraph{Batch size.}
Table~\ref{tab:vllm} reports n-gram's and \name's speedups over vLLM greedy decoding at batch~4, 8 and 16 (batch~1 is in Table~\ref{tab:ngram_compare}).
Both gains shrink as the batch grows, because a larger batch already keeps the GPU busy.
At batch~16, \name adds only 1.04--1.08$\times$ for VibeVoice-ASR on AMI, while on Mandarin, where copying through repetition loops pays at any batch size, n-gram keeps 1.65--1.76$\times$ and \name 1.30--1.48$\times$.
Accuracy is unchanged within bf16 noise at every batch size.

\begin{table}[t]
\centering
\small
\resizebox{\linewidth}{!}{%
\begin{tabular}{lllrrrr}
\toprule
Model & Batch & Method & IHM & SDM & Ali & AISH \\
\midrule
\multirow{6}{*}{VibeVoice} & \multirow{2}{*}{4} & n-gram & 1.14 & 1.15 & \textbf{4.46}$^\dagger$ & \textbf{1.26} \\
 & & \name & \textbf{1.34} & \textbf{1.26} & 4.19$^\dagger$ & 1.19 \\
\cmidrule(lr){2-7}
 & \multirow{2}{*}{8} & n-gram & 1.69 & 1.65 & \textbf{1.77} & \textbf{1.67} \\
 & & \name & \textbf{1.83} & \textbf{1.71} & 1.64 & 1.48 \\
\cmidrule(lr){2-7}
 & \multirow{2}{*}{16} & n-gram & \textbf{1.09} & 1.03 & \textbf{1.76} & \textbf{1.65} \\
 & & \name & 1.08 & \textbf{1.04} & 1.48 & 1.30 \\
\midrule
\multirow{6}{*}{MOSS} & \multirow{2}{*}{4} & n-gram & 0.81 & 0.77 & \textbf{0.81} & \textbf{0.76} \\
 & & \name & \textbf{0.93} & \textbf{0.85} & 0.77 & 0.68 \\
\cmidrule(lr){2-7}
 & \multirow{2}{*}{8} & n-gram & \textbf{0.87} & 0.87 & \textbf{0.86} & \textbf{0.80} \\
 & & \name & \textbf{0.87} & \textbf{0.88} & 0.78 & 0.57 \\
\cmidrule(lr){2-7}
 & \multirow{2}{*}{16} & n-gram & \textbf{0.96} & \textbf{0.89} & \textbf{0.94} & \textbf{0.86} \\
 & & \name & 0.75 & 0.69 & 0.68 & 0.40 \\
\bottomrule
\end{tabular}}
\caption{Speedup in decoding time over greedy decoding at the same batch size for vLLM n-gram and \name, with 330\,s chunks over the whole dataset.
Times include \name's drafter and exclude the diarization pass, which is the same for all.
Batch~1 is in Table~\ref{tab:ngram_compare}.
$^\dagger$Greedy baseline inflated by repetition loops, so absolute times are not directly comparable.}
\label{tab:vllm}
\end{table}

\paragraph{Acceptance.}
Table~\ref{tab:acceptance} reports draft acceptance at batch~8.
$\tau$ is the mean number of tokens produced per verification step (accepted draft tokens plus the model's own token), computed from vLLM's speculative-decoding counters over each dataset's run.
With loop copy, \name accepts more than n-gram for VibeVoice-ASR on AMI-IHM and AMI-SDM, and for MOSS-Transcribe-Diarize on every dataset.
For VibeVoice-ASR, n-gram stays ahead on the two Mandarin datasets, where repetition loops make up a large share of the decoded tokens and the Whisper-small drafter is weaker.

\begin{table}[t]
\centering
\small
\resizebox{\linewidth}{!}{%
\begin{tabular}{llrrrr}
\toprule
Model & Method & IHM & SDM & Ali & AISH \\
\midrule
\multirow{2}{*}{VibeVoice} & n-gram & 2.04 & 2.06 & \textbf{3.73} & \textbf{3.10} \\
 & \name & \textbf{2.31} & \textbf{2.17} & 3.28 & 2.66 \\
\midrule
\multirow{2}{*}{MOSS} & n-gram & 1.56 & 1.50 & 1.58 & 1.51 \\
 & \name & \textbf{1.87} & \textbf{1.73} & \textbf{1.66} & \textbf{1.61} \\
\bottomrule
\end{tabular}}
\caption{Mean tokens per verification step $\tau$ at batch~8 ($k=8$, 330\,s chunks, vLLM).}
\label{tab:acceptance}
\end{table}

\section{Failure Analysis}
\label{sec:appendix-failure-analysis}
We identify two failure modes that affect both chunked and single-pass inference, though with different impacts.

\paragraph{Repetition loops.}
Autoregressive decoding can collapse into repeating a token or phrase indefinitely (e.g., \texttt{"yeah ha ha ha ha ha ha ..."}), a known failure mode of sequence-to-sequence models~\cite{ahn2026whispercdaccuratelongformspeech}.
Over three single-pass runs (batch 1, 4 and 8), VibeVoice-ASR fails in 5 of 48 recording runs on AMI-IHM (10.4\%), 1/48 on AMI-SDM (2.1\%), 16/84 on AliMeeting (19.0\%) and 5/60 on AISHELL-4 (8.3\%), and no recording under 30 minutes fails.
Failures are stochastic, as the same recording can succeed in one run and fail in another.
MOSS-Transcribe-Diarize fails in 4/48 AMI-IHM and 3/48 AMI-SDM recording runs, all from recordings of 36--50 minutes, and never on Mandarin.
At 330\,s, it rarely loops (0--1 looped chunks per dataset).
We attempted to mitigate loops by varying the repetition penalty, but found no setting that reliably prevented them without degrading transcription quality on non-failing recordings.

\paragraph{Timestamp inversions.}
A second failure mode produces intervals where the predicted end time precedes the start time (e.g., \texttt{start=2633.97s, end=2633.90s}).
These inversions are detected and filtered during output stitching, but indicate that timestamp predictions are occasionally unreliable.

\section{Metric Computation}
\label{sec:appendix-metrics}
\label{sec:appendix-norm}

\paragraph{Error rates.} For a reference with $N$ words (characters for Mandarin), $\mathrm{WER} = (S + D + I)/N$, where $S$, $D$ and $I$ are the substitutions, deletions and insertions of the minimum-edit alignment between the hypothesis and the reference.
Let $R_1,\dots,R_K$ be the reference words of each speaker and $H_1,\dots,H_M$ those of each hypothesis speaker, each in time order.
Then
\begin{equation}
    \mathrm{cpWER} = \min_{\sigma} \frac{\sum_{k} E\big(R_k, H_{\sigma(k)}\big)}{\sum_k |R_k|},
\end{equation}
where $\sigma$ assigns hypothesis speakers to reference speakers one-to-one (an unmatched speaker is paired with an empty stream, so its words count as insertions or deletions) and $E$ counts the edits $S{+}D{+}I$.
tcpWER uses the same formula, but $E$ is computed under a time constraint.
Each word takes a time from its segment (MeetEval's default pseudo word-level timing), and a hypothesis word may only be aligned with a reference word whose time lies within a 5\,s collar.
Otherwise, it counts as an insertion and the reference word as a deletion.

\paragraph{Normalization.} All hypotheses and references are normalized before scoring.
We apply the following steps in order:
\begin{enumerate}
  \item \textbf{Unicode normalization.} NFKC normalization, which maps full-width characters to their half-width forms.
  \item \textbf{Lowercasing.} All characters are converted to lowercase.
  \item \textbf{Non-speech removal.} Bracketed non-speech labels and their contents are removed (e.g., \texttt{[Human Sounds]}, \texttt{[Laughter]} produced by VibeVoice-ASR, and \texttt{<noise>}-style reference annotations).
      Removing them before punctuation keeps their contents from being scored as words.
  \item \textbf{Filler removal.} English filled pauses (\emph{uh}, \emph{um}, \emph{mm}, \emph{hm}, \emph{hmm}, \emph{mhm}) are deleted.
      Mandarin fillers are not removed.
  \item \textbf{Punctuation removal.} All remaining punctuation is stripped.
\end{enumerate}
For Mandarin, we score characters (CER).
A space is inserted between CJK characters before diffing, so word-level scorers, including cpWER and tcpWER, operate on characters, while Latin and digit runs stay single tokens.
WhisperX output is converted from traditional to simplified characters.
Normalization is applied identically to both the hypothesis and the reference transcripts to ensure comparability.

\end{document}